\documentclass[poms,final,nonblindrev]{poms1_V1} 
\usepackage{comment}
\usepackage{subfigure, epsfig}
\usepackage{natbib}
\usepackage{amsmath}
\usepackage{hyperref} 
 \bibpunct[, ]{(}{)}{,}{a}{}{,}%
 \def\bibfont{\small}%

\TheoremsNumberedThrough     

\ECRepeatTheorems

\EquationsNumberedThrough    
\begin{document}


\RUNAUTHOR{Bhattacharya and Kumar}

\RUNTITLE{Optimizing delivery for quick commerce factoring qualitative assessment of generated routes}

\TITLE{Optimizing delivery for quick commerce factoring qualitative assessment of generated routes}


\ARTICLEAUTHORS{%
\AUTHOR{Milon Bhattacharya}
\AFF{Dept. of Production \& Operations Management, IIM Visakhapatnam, \EMAIL{milon.bhattacharya25-08@iimv.ac.in}} 
\AUTHOR{Milan Kumar}
\AFF{Dept. of Production \& Operations Management, IIM Visakhapatnam, \EMAIL{milan@iimv.ac.in}}

}

\ABSTRACT{%
India’s e-commerce market is projected to grow rapidly, with last-mile delivery accounting for nearly half of operational expenses. Although vehicle routing problem (VRP) based solvers are widely used for delivery planning, their effectiveness in real-world scenarios is limited due to unstructured addresses, incomplete maps, and computational constraints in distance estimation. This study proposes a framework that employs large language models (LLMs) to critique VRP-generated routes against policy-based criterias, allowing logistics operators to evaluate and prioritize more efficient delivery plans. As a illustation of our approach we generate, annotate and evaluated 400 cases using large language models. Our study found that open-source LLMs identified routing issues with \textcolor{red}{79\%} accuracy, while proprietary reasoning models achieved reach upto \textcolor{red}{86\%}. The results demonstrate that LLM-based evaluation of VRP-generated routes can be an effective and scalable layer of evaluation which goes beyond beyond conventional distance and time based metrics. This has implications for improving cost efficiency, delivery reliability, and sustainability in last-mile logistics, especially for developing countries like India.
}

\maketitle
\section{Introduction}
The e-commerce industry in India is expected to grow at a CAGR of 45 percent in the next decade and reach a market size of approximately 400 billion \cite{IBEF2025}. The backbone of the industry is the delivery process, especially the last mile, which constitutes 41 – 53 percent of the total revenue expenditure for any e-commerce firm \cite{Maersk2025}. Efficiency gains at this step however small, would lead to substantial cost savings for such e-commerce operators. In addition, as the market grows, concerns related to sustainability and profitable growth demand that the delivery of goods be more resource efficient.

\section{Background}
Last-mile order delivery is an achillis heel for most e-commerce providers, especially firms which handle large volumes of orders \cite{macioszek2017first}. These orders consist of low-volume/low-value items which mostly contains food and groceries. Such orders do not require large delivery capacity of vehicles and allow grouping delivery of orders \cite{cleophas2014deliveries}. 
This sector is unorganized, mired in regulation and very geography-specific \cite{ehrler2021challenges} \cite{archetti2021recent}. In the Indian context, most of the e-commerce companies outsource last-mile delivery services to third-party logistics players and prefer either to operate as a marketplace or as an \emph{asset-lite} operator \cite{raman2021comprehensive}. 

These players are MSMEs who cater to a very specific area and service most of the e-commerce provider (both big and small) delivering to that geography \cite{puram2022last}. Typically, they would collate orders from all such providers and deliver them as one or more routes. The problem with this practice is that such providers must service orders which may be scattered across the geography without the scope of efficient grouping of these deliveries. Also, the delivery locations for many orders are not correctly identifiable due to incorrect address and lack of map coverage of points-of-interests (PoI) \cite{panigrahi2016commerce}. Furthermore, the large volume of orders prohibits exhaustive pair-wise distance calculation, which in turn require approximations for finding the distances between each drop/pickup. This is referred to as the \emph{distance matrix} in VRP parlance. This matrix is critical for solving any version of the vehicle routing problem. The optimization of an approximate distance matrix without the considering the actual route leads to inefficient last-mile deliveries, where the rider might be required to travel additional distances or perform risky driving behaviours (e.g. driving in a no-vehicle road)

\section{Problem Statement}
Our attempt in this study has been to develop a framework which could identify problems in the routes generated during the VRP process. Specifically, we use large foundation model (LLM) to review generated routes against a suer specified policy. Such a review would allow the logistics operator to rate, select and prioritize better routes while planning for deliveries. The policy can be specified using natural language, unlike the VRP solvers where constraints have to be specified as a closed form expression.  This allows users to check for varied conditions / senarios. The use of LLMs allows inclusion of additional third party tools e.g. APIs for retrieving weather or topography.
Such an assessment would be very relevant for for e-commerce applications where the number of orders to be delivered in a slot are large and the time available to create, select and assign routes from these orders is not realtime (in the range of a few hours). For such scenarios, the quality of the generated route is of primary importance. 

\section{Research objectives}
This study attempts to investigate the following aspects in relation to the practice of last-mile delivery in India. namely,  

\begin{itemize}
    \item \emph{RQ1}: How do solutions from popular VRP tools work in real-world scenarios? 
    \item \emph{RQ2}: If there are problems with such solutions, is it possible to detect those in an automated fashion. 
    \item \emph{RQ3}: is it possible to use large language models to automate such evaluations. 
    \item \emph{RQ4}: Finally, how useful are these evaluations and if they can make a business impact. 

\end{itemize} 

\section{Literature review}
The vehicle routing problem is a well-studied problem with multiple solution approaches \cite{Toth2002}. These approaches can be roughly classified into three main categories: exact methods where the problem is solved using known mathematical constructs like convex optimization \cite{Liu2023a}, exact methods with heuristics that attempt to reduce the solution search space \cite{Sidorov2021} and metaheuristics \cite{Labadie2016} which attempt to build models which can learn to generate and improve routes iteratively. the exact methods are not scalable due to the intractable nature of the solution, hence the other two approaches have received the maximum attention. Both have been shown to improve performance when compared using metrics like total distance traveled or total time to deliver.   
\break
In the last decade, the availability of large deep learning models has led to their usage in the field of supply chain management  \cite{Dhara2023} \cite{Li2023} \cite{Jannelli2024} . One such application is to generate vehicle routing solutions.
Such applications can roughly be categorized either as those which use the LLM as optimizers (also known as neural solvers) for improving the routing solution \cite{Wu2024} \cite{Yang2023} \cite{Cao2025} \cite{Huang2024} \cite{Thanh2025} \cite{Shi2019}, or as approaches which use LLMs to express the constraints while can be fed then to a conventional VRP solver \cite{Jiang2025}. \cite{bogyrbayeva2024machine} contains a comprehensive survey of the taxonomy of such approaches in respect to their learning paradigms, solution structures, underlying models, and algorithms. Most of the approaches use a deep learning architecture (encoder-decoder transformer \cite{bertoa2025foundation} , MoE \cite{Zhou2024b}) to train on the synthetic VRP problems and their solutions from conventional solvers.
\break
Other attempts have used LLMs as a driving assistant for providing real-time updates \cite{Zhou2024}. \cite{Ulmer2017} is a survey of such approaches which use LLMs as reasoning tools during real-time navigation. \cite{Ye2024} uses the generate-reflect-correct paradigm from the LLM reasoning literature \cite{xiao2025foundations} and attempts to iteratively improve the VRP solution. \cite{Huang2024} uses a textual description of the VRP problems as a prompt and uses the LLM response to create a routing order. All such attempts have limited themselves to the route generation (or improvement process). Most of the studies have quoted efficiency gains on artificial benchmarks like \href{https://vrp.atd-lab.inf.puc-rio.br/index.php/en/}{CVRPLIB} or synthetically generated datasets. However, no study attempted using LLM for qualitative evaluations for VRP use-cases. \cite{huang2025multimodal} is a study which is comes closest to our work. It attempts to enhance the optimization performance using multimodal LLM capable of processing both textual and visual prompts for deeper insights of the processed optimization problem. Another study utilizes LLM agents to generate a solution analysis function for the contraints in VRP which lead to an infeasible solution \cite{Li2025b}. \cite{Liu2023b} attempts to use the historical data of completed rides and the implicit knowledge of the rider to improve/guide navigation for the current route. Finally \cite{Kikuta2024} uses a LLM to explain the routing order.
However, all these attempts limit themselves to the route generation process.

\section{Methodology}
As an illustration of the problem and our attempt to solve it, we walk through the steps of a typical route generation process. To mimic real-world delivery demand scenarios, a dataset was created which corresponded to real-world delivery scenarios. The OpenStreetMap (OSM)\cite{bennett2010openstreetmap} data for the city of Bengaluru was used as a study area. This dataset was labeled for the presence / absence of three most common problems encountered by drivers. Further, to simulate route generation, we generate test solutions for these locations using the OR Tools package. Finally, we propose a qualitative approach for critiquing VRP solutions and demonstrate the approach on a sample dataset using textbf{five} open-source and one proprietary language models. As a final step, we compare these open-source models for their ability to reason and identify problems and compare the results with the latest proprietary vision model. To allow replicability, we share the samples, results and the accompanying code on 
\href{https://github.com/bmilon/POMS2025}{GitHub}.
\subsection{Study Area}
To sample real-world routes, it was decided to use map data for an actual city. The city of Bengaluru was selected considering the complexity and variety in the topology of the routes.  This data includes the road network, water bodies and other attributes, including the base image. The base image is an image based representation of all the overlaying features.

\begin{figure}[!ht]
        \centering
    \caption{Bengaluru from OpenStreetMaps}
 \includegraphics[width=7cm]{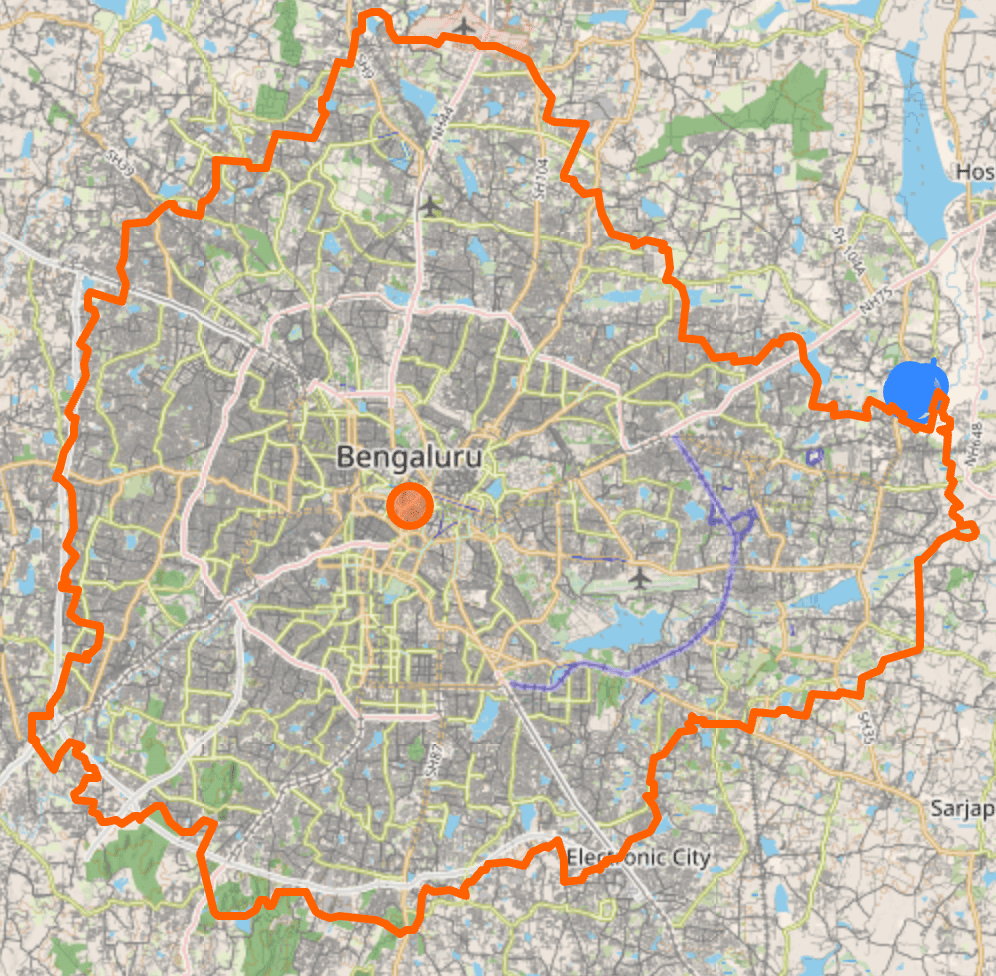}       
 \label{fig:osmblr}
\end{figure}

\subsection{Data generation process}
Once the study area was finalized, the next step was to select possible locations of deliveries, including the depot from where the deliveries had to be fulfilled. For simplicity sake, the total demand across all delivery points  $D_1 , D_2 .... D_N $ is less than carrying capacity of the vehicle.  
\begin{equation}
\sum_{i=1}^{N} D_i \ll C_{vehicle}
\end{equation}
The map retrieved from OSM is in the form of \emph{LineString} objects. These can be converted into a set of points, grouped as cluster and sampled using a two-stage process.

\begin{equation}
\text{Clustering:} \quad U = \bigcup_{k=1}^{K} C_k, \quad C_i \cap C_j = \varnothing \;\; (i \neq j) \\[6pt] \text{where $C_k'$ is the sample from cluster $C_k$}
\end{equation}

\begin{equation}
 \text{Cluster selection:} \quad S \subseteq \{1,2,\dots,K\}, \quad |S| = m \\[6pt]; \quad \text{where $K$ is the number of derived clusters}
\end{equation}

\begin{equation}
\text{Sample size (stage-I)):} \quad n = \sum_{k \in S} |C_k| \\[6pt]
\end{equation}

\begin{equation}
 \text{Sample size (stage-II)):} \quad n = \sum_{k \in S} n_k, \quad n_k \leq |C_k| \\[6pt]
\end{equation}

For each of the experiments $K \approx 20{,}000$ and  $10 \leq n \leq 20$ which is typically the order volume in each service area (SA).

\subsection{Route Generation}
As described in  \cite{Toth2002} the vehicle routing problem can be described as follows,

\begin{equation}
\begin{aligned}
&\textbf{Sets:} \\
&c_{ij} && \text{travel cost from node $i$ to $j$} \\
&d_i   && \text{demand of customer $i$} \\
&Q     && \text{vehicle capacity} \\[6pt]
\end{aligned}
\end{equation}

\begin{equation}
\begin{aligned}
&\textbf{Parameters:} \\
&c_{ij} && \text{travel cost from node $i$ to $j$} \\
&d_i   && \text{demand of customer $i$} \\
&Q     && \text{vehicle capacity} \\[6pt]
\end{aligned}
\end{equation}

\begin{equation}
\begin{aligned}
    &\textbf{Decision variables:} \\
    &x_{ijk} = 
    \begin{cases}
    1 & \text{if vehicle $k$ travels from $i$ to $j$} \\
    0 & \text{otherwise}
    \end{cases} \\[6pt]    
\end{aligned}
\end{equation}

\begin{equation}
\begin{aligned}
&\textbf{Objective:} \\
&\min \sum_{k \in K} \sum_{i \in V} \sum_{j \in V} c_{ij} \, x_{ijk} \\[10pt]
\end{aligned}
\end{equation}

\begin{equation}
\begin{aligned}
&\textbf{Constraints:} \\[4pt]
&\sum_{k \in K} \sum_{j \in V} x_{ijk} = 1, 
&& \forall i \in N 
\quad \text{(each customer visited exactly once)} \\[6pt]
\end{aligned}
\end{equation}

\begin{equation}
\begin{aligned}
&\sum_{j \in V} x_{0jk} = 1, 
\quad \sum_{i \in V} x_{i0k} = 1, 
&& \forall k \in K 
\quad \text{(each vehicle leaves and returns to depot)} \\[6pt]
\end{aligned}
\end{equation}

\begin{equation}
\begin{aligned}
&\sum_{i \in V} \sum_{j \in V} d_j \, x_{ijk} \leq Q, 
&& \forall k \in K 
\quad \text{(vehicle capacity constraint)} \\[6pt]
\end{aligned}
\end{equation}

\begin{equation}
\begin{aligned}
&x_{ijk} \in \{0,1\}, 
&& \forall i,j \in V, \; k \in K
\quad \text{(Sub-tour elimination)} \\[6pt]
\end{aligned}
\end{equation}

The exact solution to the VRP is difficult as the Vehicle Routing Problem is NP-hard, since it generalizes the Traveling Salesman Problem (TSP). \cite{dantzig1959vrp}. As mentioned before, different approaches like branch and cut \cite{brunetta1997branch} , branch and price \cite{barnhart1998branch} have been used to obtain a \emph{near-optimal} solution. Google OR Tools \cite{ortools_routing} is a popular implementation of such approaches. It also uses some soft computing techniques to improve the solution further. Figure~\ref{fig:sampleroute} is an illustration of the routing order from one of the experiments.

For each of the experiments, a routing order was generated using OR Tools approach. A total of 400 routes are considered for this exercise. The locations along with the OSM basemap are available within our github repository.

\subsection{Generation of driving directions}
\begin{figure}[!ht]
    \centering
    \begin{minipage}{.5\textwidth}
        \centering
    \caption{Routing order from OR Tools}
\includegraphics[width=0.8\textwidth]{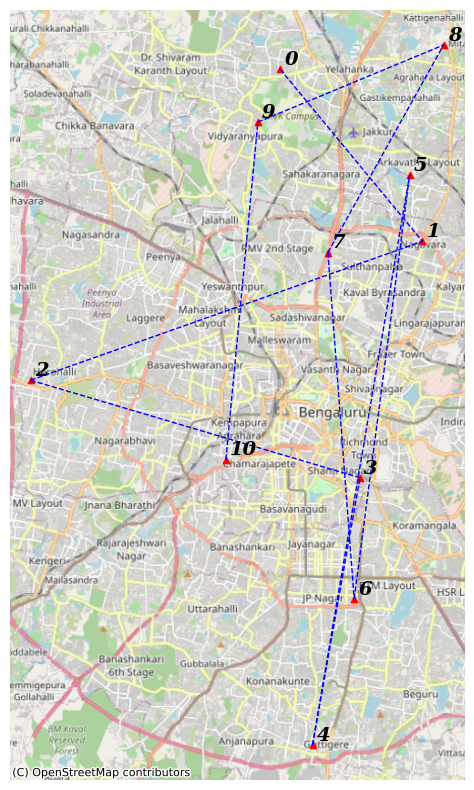}
 \label{fig:sampleroute}  
     \end{minipage}%
    \begin{minipage}{.5\textwidth}
        \centering
    \caption{Actual Driving directions}
\includegraphics[width=0.8\textwidth]{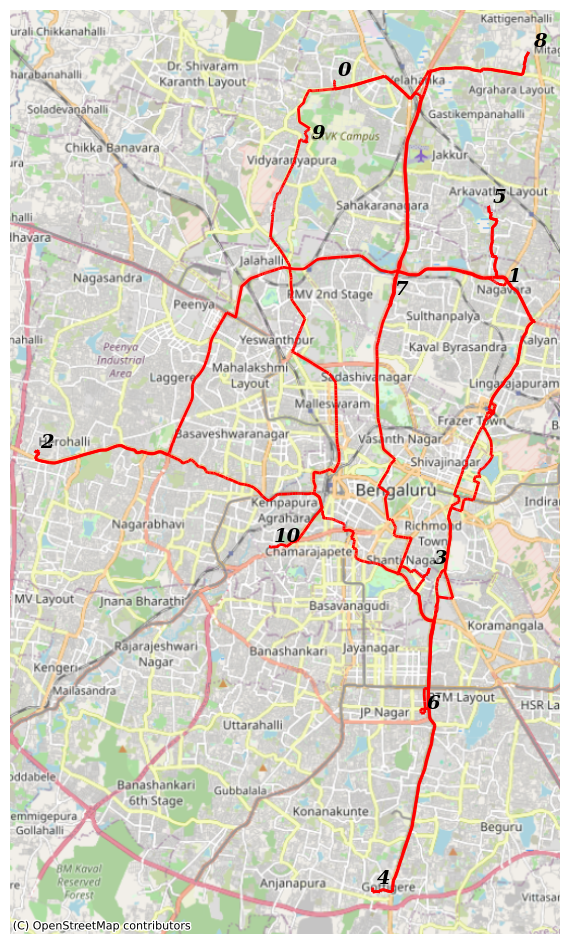}     
\label{fig:ortools_routing}
    \end{minipage}
    \vspace*{10pt}
\end{figure}

Once the delivering order is decided, the travel paths are generated. 
To allow us to share data publicly and aid reproducibility, we use the in-built 
\href{https://osmnx.readthedocs.io/en/stable/}{OSMX} library for generating the actual driving directions as shown in Figure~\ref{fig:ortools_routing}.

\subsection{Qualitative route evaluation}
For performing the evaluations, each leg of the route was individually evaluated using the LLM. To ensure that the output
token budget is not crossed, the ability of the language model was tested using the following four
questions namely, 

\begin{question} \label{question:Q1}
    Does any of the routes cross a water body ?
\end{question}

\begin{question} \label{question:Q2}
    Does it pass through a railway-crossing / railway line ?
\end{question}

\begin{question} \label{question:Q3}
    Does it pass through pedestrian area ?
\end{question}

\begin{question} \label{question:Q4}
    Does the route pass through a park or forested area ?
\end{question}

These questions are part of a larger set of issues encountered by the delivery riders. They were uncovered using a grounded theory based survey among twenty delivery 
in and around Bengaluru. The survey highlighted operational realities of last-mile delivery and common challenges encountered during last-mile deliveries by delivery personnel. 
The discussions with such riders would be published as a seperate qualitative study. The exact 
prompt which encodes these questions is available as a seperate function in the github repository.

\subsection{Labelling of the dataset}
\begin{figure}[!ht]
        \centering
    \caption{Sample image from the generated dataset}
 \includegraphics[width=10cm]{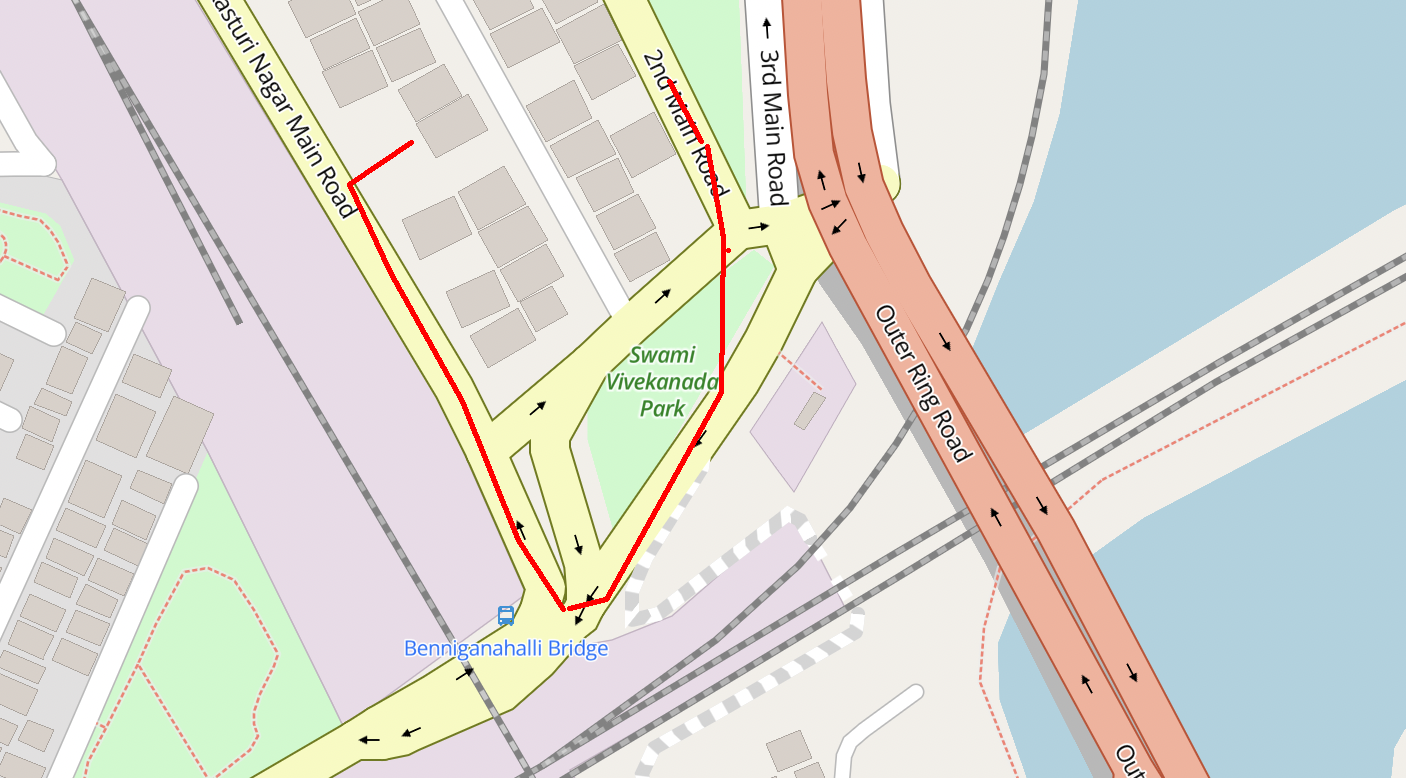}       
 \label{fig:sample_image_from_dataset}
\end{figure}

One of the aims of the study is to create a benchmark dataset which can be used for subsequent studies on the effectiveness 
of foundation models in studying vehicle routing problems. The use of open source datasets and tools along with
publicly availabile of the codebase base would a step in this direction. 
To correctly evaluate the results from the language model, the images were labelled against each of the questions listed 
in this study. For example the image in Figure~\ref{fig:sample_image_from_dataset} the annotators response is given in  
table \ref{table:sample_annotation}. For coordinating between the volunteers the open source tool \href{https://labelstud.io/}{label-studio} was used.
Finally, the protocols and best practices followed for this exercise were inspired from \cite{liao2021towards}.
Additional images, along with the annotations of the images,
 The complete dataset created for this study is available 
 \href{https://iimv1-my.sharepoint.com/:u:/g/personal/milon_bhattacharya25-08_iimv_ac_in/EYlWCl66-2lPqyAvguwMLC8BNl0GKppr-T5JaFHY3piVxQ?e=o14lmj}{here} for download.

\begin{figure}[!ht]
        \centering
    \caption{Label Studio}
 \includegraphics[width=15cm]{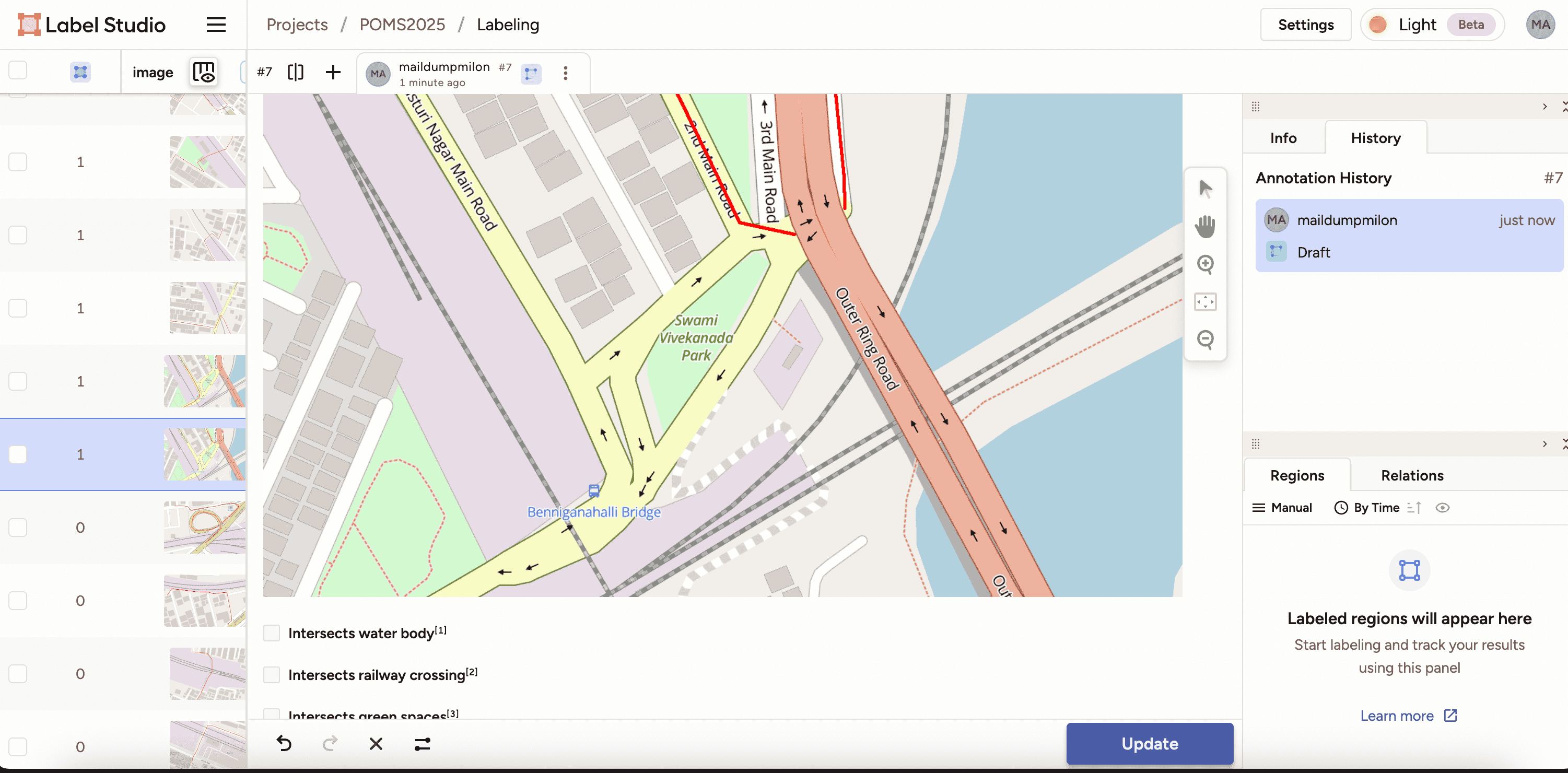}       
 \label{fig:labelstudioui}
\end{figure}

\begin{table}[!ht]
\caption{Sample annotation for Figure~\ref{fig:sample_image_from_dataset} }
\centering
        {\def\arraystretch{1}  
\begin{tabular*}{0.5\textwidth}{@{\extracolsep{\fill}}lcc}
\hline
\hline

Question & Answer \\
\hline
crosses a water body       & No   \\
passes through a railway-crossing       & No   \\
goes through a pedestrian area     & No   \\
Does it cross a park or forest area     & Yes   \\
\hline
\hline
\end{tabular*}
\label{table:sample_annotation}
}
\end{table}

\subsection{Result compilation}
The multi-modal language models output a 
raw string as a evaluation of the image.
 Due to difference in training approaches, each model may either output unformatted text, JSON or markdown. To allow consistent evaluation,
a second language model \href{https://huggingface.co/mistralai/Mixtral-8x7B-v0.1}{mistralai/Mixtral-8x7B-v0.1} \cite{jiang2024mixtral} was 
used to extract the relevant information from the output of the vision model. \break The results were tabulated 
and were reviewed by the authors and an external reviewer. The attempt of the review was to check if the 
answers provided by the language model were same as the actual label. In our experience, not more than 2\% of the responses were summarized incorrectly.

\section{Results and Discussion}
The study showed that given a budget in terms of compute and time, it is possible to correctly identify different types of problems within a route.  For the 400 examples created and manually annotated for this study, the best performing open-source model \emph{gemma3} was able to identify the problems for 79 percent of the cases. The accuracy went up to 86 percent when proprietary model was used with a larger token budget for reasoning. 

\begin{table}[!ht]
\caption{Accuracy versus inference Time}
\centering
        {\def\arraystretch{1.6}  
\begin{tabular*}{0.8\textwidth}{@{\extracolsep{\fill}}lccc}
\hline
\hline
Model Name & Accuracy  &  Average Inference Time (Seconds) &  Model Size  \\
  &   &   ($ \mu \pm 1\sigma $) & (Billion) \\
\hline 
gemma3:12b	        & \textcolor{green}{0.79}    & 38.40  &  	12 \\
mistral-small3.2    &  0.62   & 94.15  &  	24\\
qwen2.5vl:7b        &  0.59   & 65.44  &  	7\\
minicpm-v           &  0.42   & 36.74  &    8  \\
llama3.2-vision     &  0.41   & 116.82 &  	11\\
granite3.2-vision   &  0.34   & 54.65  &  	2 \\

GPT 5               &  0.86   & \textcolor{red}{60} &  	\textcolor{red}{1760} \\
\hline
\hline
\end{tabular*}
\label{table:compare_models1}
}
\end{table}

\subsubsection{How do solutions from popular VRP tools work in real-world scenarios}

The VRP tools, while mathematically sound, often fall short in real-world scenarios due to several factors.
The reasons include the prevalence of unstructured addresses, incomplete or outdated map data, and the use of approximations for computing 
the pair-wise distance matrix. This is especially true when  distance estimations are required 
across a large number of delivery points. 
This often leads to routes that are theoretically optimal but practically inefficient, 
forcing delivery personnel to make on-the-fly adjustments that increase delivery times and costs. In our experiments, the use of 
line / \emph{crow-flying} distance was used to create the distance matrix. The resulting order fulfillment schedule looked in-efficient, when 
reviewed by human annotators. 

\subsubsection{Is it possible to detect problems in an automated fashion}
In the context of the limited scope of this study, we can conclude that it is possible to automate such evaluations
completely using open source tools and frameworks. 

\subsubsection{Is it possible to use large language models to automate such evaluations}
In our study, the models used for evaluation were open source in nature. Also, the version of the model used was the smallest in the individual
model family and optimized for on-device inference. For example we used Llama Scout from the \href{https://ai.meta.com/blog/llama-4-multimodal-intelligence/}{Llama} 
family of models released by Meta, which is a 17B active parameter model. It's model size is only 6\% of the best model in the 
family and yet it exceeds the performance of previous model on visual reasoning tasks e.g. ChartQA  \cite{masry2022chartqabenchmarkquestionanswering}. As is 
evident from Table~\ref{table:compare_models1} and Table~\ref{table:compare_models2}, it is possible to use \emph{off-the-shelf} models on commodity hardware to 
get upwards of 70 percent detection rate. For this study a Linux device running on AMD EPYC 7402 24-Core Processor was used. The models were hosted locally using Ollama \cite{marcondes2025using} with maximum usage of 25 GB of GPU memory (on a NVIDIA RTX A6000).

 \section{Limitations}

 The study demonstrates the ability of performing qualitative evaluations of last-mile delivery routes in an automated fashion. However,
 a couple of problems were observed during our assessment. 

 \subsubsection{Generic models}
\begin{figure}[!ht]
        \centering
    \caption{Different symbols used for railway line}
 \includegraphics[width=10cm]{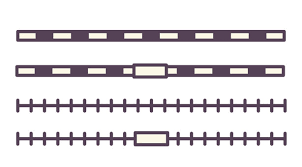}       
 \label{fig:railway_symbol}
\end{figure}

The vision-models from each family have been training on generic image understanding datasets
 e.g. Journey DB, MMIU Chart QA etc. The amount of geospatial data and related tasks in such datasets are limited if not absent. Hence, the models 
 tend to get confused when they encounter very specific map symbology. e.g. railway lines can be either represented
 as a double line or a single hatched line as shown in Figure~\ref{fig:railway_symbol}. A fintuned model is expected to work better at identifying 
 such symbology, especially when the basemap resolution is poor.

 \subsubsection{Quantization error}
\begin{figure}[!ht] 
        \centering
    \caption{Example which confuses smaller models}
 \includegraphics[width=10cm]{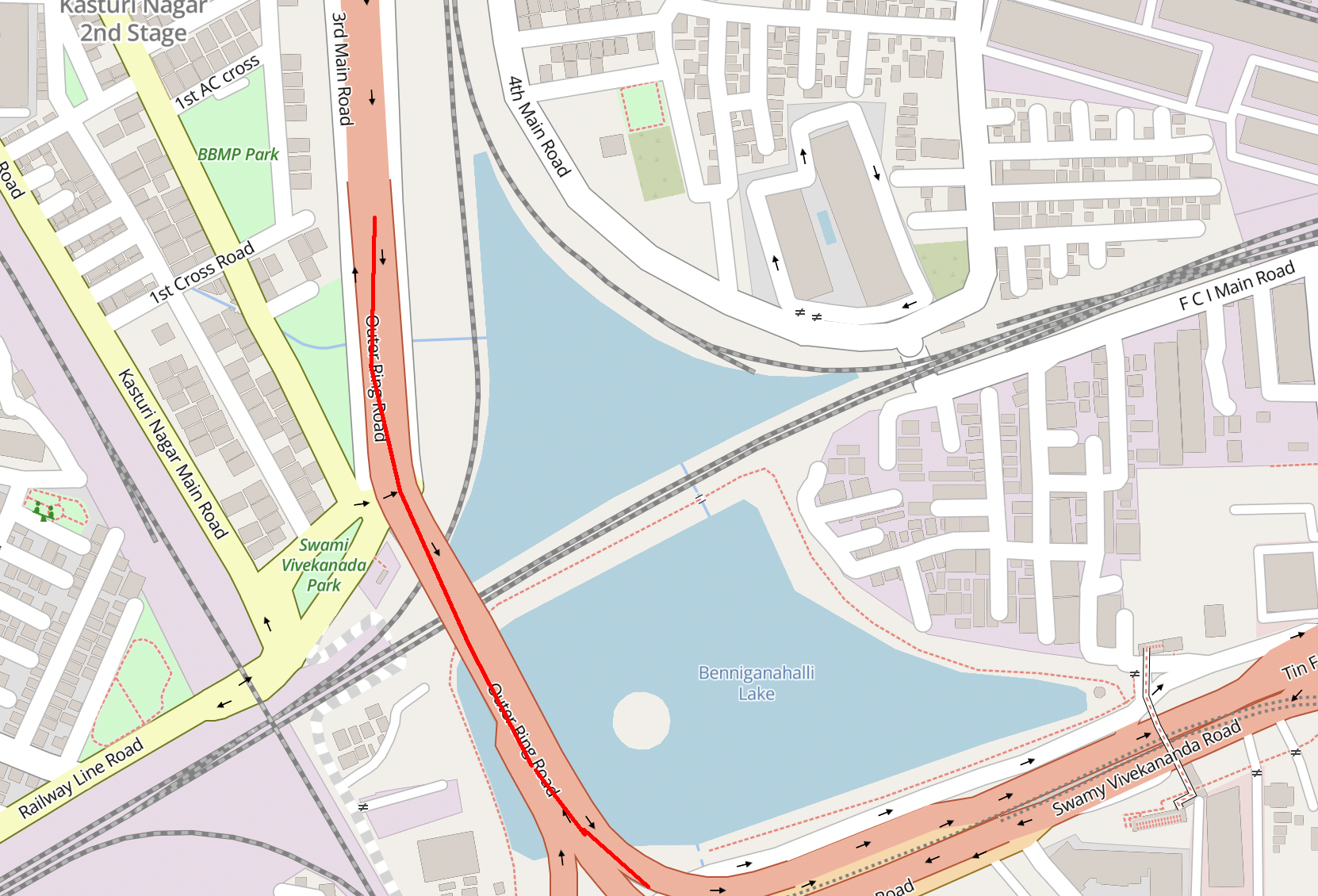}       
 \label{fig:confusing_image}
\end{figure}

The vision models work on the concept of image embeddings, where the input image is converted into a series of tiles, which are converted into vectors \cite{ramachandran2019stand}. Based on the kind of fusion (early versus late) \cite{huang2024early}, the model may chose to perform cross-attention or feed the embedding with the text embeddings. For this study the smallest of each of the family of models were selected to allow to local inference without the need to query a hosted model. Such models are have a parameter count smaller than the original model and are either quantized or distilled or both. The loss of  performance (when measured on standard benhmarks) is one of the known problems with such models. \cite{gou2021knowledge}. Also, larger models have been found to be more sample efficient \cite{zhai2022scaling}.
 In, our study we found that the model's ability to differntiate between closly spaced features reduces as model size decreases. For example for the image in Figure~\ref{fig:confusing_image} confuses smaller models like \emph{mistral-small3.2} which incorrectly concludes that the marked route (in red) intersects a railway line. A larger model like \emph{GPT5} (estimated to have 1.5T active parameters ) was able to reason the spatial seperation between the route and the underlying railway line. There is no straightjacket rule for deciding the most apt model for visual reasoning. However, \cite{shi2024we} suggested use of multi-scale inference of smaller models to match the performance of larger models. In subsequent studies we would attempt to quantify this trade-off between model size and performance.

\begin{table}[!ht]
\caption{Results obtained using different language models (TPR versus FPR)}
\centering
        {\def\arraystretch{1.6}  
\begin{tabular*}{0.7\textwidth}{@{\extracolsep{\fill}}lccccc}

\hline
\hline
Sr. No. & Model Name & TPR & FPR & Precision & Recall   \\
\hline
1. & gemma3:12b            & 0.70 & 0.22 & 0.91 & 0.69 \\
2. & mistral-small3.2      & 0.50 & 0.04 & 0.96 & 0.5 \\
3. & qwen2.5vl:7b.         & 0.51 & 0.12 & 0.93 & 0.51 \\
4. & minicpm-v             & 0.23 & 0.05 & 0.91 & 0.23 \\
5. & granite3.2-vision     & 0.23 & 0 & 1.0 & 0.23 \\
6. & llama3.2-vision       & 0.15 & 0.1 & 0.88 & 0.15 \\
7. & gpt-5      &  \textcolor{red}{0.78} & 0.01 & 1.0 & 0.9 \\

\hline
\hline
\end{tabular*}
\label{table:compare_models2}
}
\end{table}

 \subsubsection{Time for inference}
\begin{figure}[!ht]
        \centering
    \caption{Accuracy versus model size and inference time}
 \includegraphics[width=12cm]{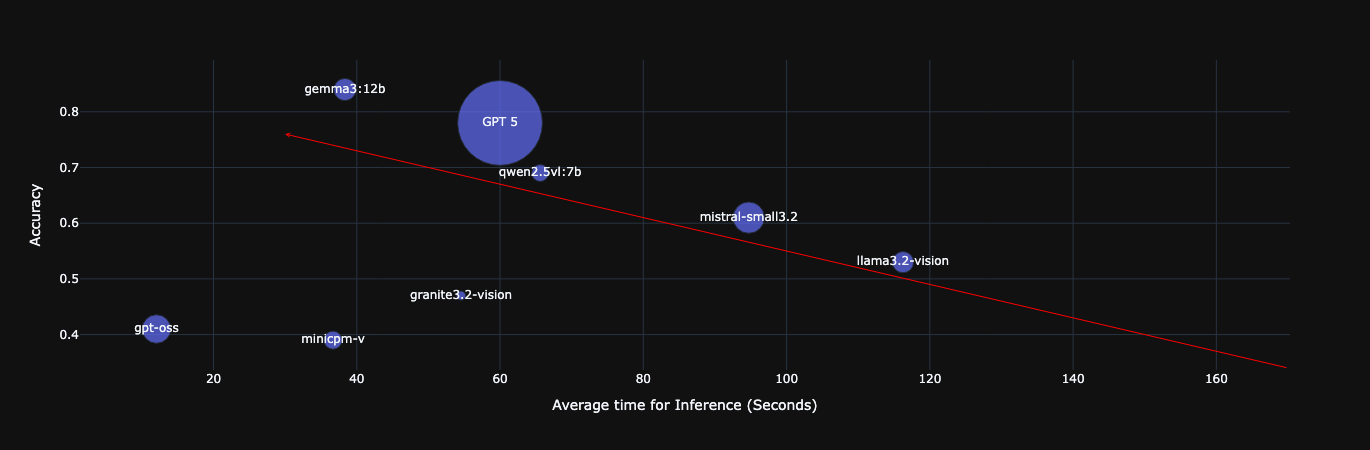}       
 \label{fig:model_size_versus_accuracy}
\end{figure}

As is evident from the Table ~\ref{table:compare_models1}, inferencing even the smallest models on from each family cannot be done in real-time mode by only using commodity hardware. It is only possible if we use specialized hardware. As is evident from Figure ~\ref{fig:model_size_versus_accuracy}, models with higher inference time tend to produce higher accuracy. However, the correspondance of model size with its performance on our task (visual reasoning) does not show a clear trend. Therefore, there is clear opportunity to optimize the selection of such models for their relevance to visual reasoning, especially those relating to geospatial data. If time is not a constraint the model could have been asked to answer questions one at a time, instead of our approach of grouping all questions in a single request. Similarly, if the resolution of the images could be increased, the chances of correctly identifying the problem (if it exists) would be much higher.

\subsubsection{Lack of consistent output}
Due to the nature of their training process , language models have to be instructed to explicitly respond in a given format \cite{ma2025should}. However, inspite of such instructions the models don't always respond consistently. This variation could be in the length of the response, the output format which could be free text, markdown or JSON, or even the choice of language. This study attempt to ameliorate such variations using a library called pydantic, which allows specification of models of expected response. In addition, the evaluation of the model response itself is done using a response-critic model where the LLM responses are compared to human responses using another LLM. This allows the evaluations to be scalable. In real-world scenarios, such comparisions would be performed by a human operator who would manually review the LLM response and verify the created routes.

 \subsubsection{Controlling the inference parameters}
In addition to choice of model family and the specific size (parameter count), other parameters that can be varied are the inference temprature and sampling methodology. \cite{li2024llm} contains a description of the recent advances in the field and can be referred for additional details on the topic. As a part of the study, a couple of experiments were conducted by varying the \emph{temperature} and the \emph{topk} parameter in the Ollama API. The changes in temperature did not lead to additional improvements in the metrics and therefore those results are not included in the final report. As is evident from the Table~\ref{table:compare_models1}, the inference time does not depend on the model size / active parameters but it also is a function of the model's ability to output succinct output. For example, \emph{llama3.2-vision}, with a parameter count of 11 Billion active parameters takes a more than twice the time to give an answer as taken by \emph{gemma3:12b} and still underperforms. Hence, controlling for output token budget may not be the best approach.

\section{Contribution to industry}
Use of routing solutions for planning delivery of goods and services is pervasive in modern industries. Today e-commerce applications are delivering courier, groceries, medicine and other items of daily use. The approach if utilized for analysing routes would be useful for evaluating the quality of routes. Not only would it help in estimating time and effort, which in turn could be used to develop better compensation mechanisms for the operators of such delivery vehicles but also would help quantify business risks for such companies. 

\section{Future Work}
During our study we identified the following areas which could have been invstigated to improve our results even better.

\section{Conclusion}
This study addressed the critical challenge of last-mile delivery optimization in the rapidly growing Indian e-commerce sector, where routes generated by conventional VRP based approaches
often produce routes that are suboptimal in practice. We proposed and validated a novel framework that leverages Large Language Models (LLMs) to perform a qualitative critique of VRP-generated routes, moving beyond traditional metrics of distance and time. Our experiments, conducted on a synthetic dataset, demonstrated the effectiveness of this approach.
 We found that both open-source and proprietary LLMs can identify practical routing issues—such as traversing parks or
  water bodies with high accuracy. This automated evaluation provides logistics operators with a 
  practical tool to evaluate and prioritize routes in an automated fashion. 
This ensures that the routes are not only mathematically optimal but also operationally feasible and efficient.
The implications for the industry are significant. By integrating this qualitative assessment layer, companies can enhance delivery reliability, 
reduce operational costs, and improve sustainability. Furthermore, it opens up avenues for creating more equitable compensation models for 
delivery personnel by accounting for real-world route complexity.While this study establishes a strong proof-of-concept, 
future research could expand on this framework by incorporating a wider range of qualitative criteria, 
testing with real-time operational data, and exploring methods to use LLM feedback to directly refine and improve the route generation process itself. 
In conclusion, our work highlights the transformative potential of LLMs in bridging the gap between abstract optimization and the complex, dynamic reality of last-mile logistics.

\ACKNOWLEDGMENT{The authors gratefully thank the reviewers Subhayan Saha and Adeitia B. PhD students from IIM Visakhapatnam for their help in annotating and reviewing the dataset.}


\bibliographystyle{pomsref}

 \let\oldbibliography\thebibliography
 \renewcommand{\thebibliography}[1]{%
    \oldbibliography{#1}%
    \baselineskip14pt 
    \setlength{\itemsep}{10pt}
 }
\bibliography{references.bib}

\begin{thebibliography}{50}
\expandafter\ifx\csname natexlab\endcsname\relax\def\natexlab#1{#1}\fi
\expandafter\ifx\csname url\endcsname\relax
  \def\url#1{{\tt #1}}\fi
\expandafter\ifx\csname urlprefix\endcsname\relax\def\urlprefix{URL }\fi
\expandafter\ifx\csname urlstyle\endcsname\relax
  \expandafter\ifx\csname doi\endcsname\relax
  \def\doi#1{doi:\discretionary{}{}{}#1}\fi \else
  \expandafter\ifx\csname doi\endcsname\relax
  \def\doi{doi:\discretionary{}{}{}\begingroup \urlstyle{rm}\Url}\fi \fi

\bibitem[{Archetti and Bertazzi(2021)}]{archetti2021recent}
Archetti, Claudia, Luca Bertazzi. 2021.
\newblock Recent challenges in routing and inventory routing: E-commerce and last-mile delivery.
\newblock {\it Networks\/}, { 77} (2), 255-268.

\bibitem[{Barnhart et~al.(1998)Barnhart, Johnson, Nemhauser, Savelsbergh, and Vance}]{barnhart1998branch}
Barnhart, Cynthia, Ellis~L Johnson, George~L Nemhauser, Martin~WP Savelsbergh, Pamela~H Vance. 1998.
\newblock Branch-and-price: Column generation for solving huge integer programs.
\newblock {\it Operations research\/}, { 46} (3), 316-329.

\bibitem[{Bennett(2010)}]{bennett2010openstreetmap}
Bennett, Jonathan. 2010.
\newblock {\it OpenStreetMap\/}.
\newblock Packt Publishing Ltd.

\bibitem[{Bertoa et~al.(2025)Bertoa, Huaa, Zepedab, Hottungb, Woudac, Land, Parka, Tierneyb, and Parka}]{bertoa2025foundation}
Bertoa, Federico, Chuanbo Huaa, Nayeli~Gast Zepedab, Andr{\'e} Hottungb, Niels Woudac, Leon Land, Junyoung Parka, Kevin Tierneyb, Jinkyoo Parka. 2025.
\newblock A foundation model for vehicle routing problems, .

\bibitem[{Bogyrbayeva et~al.(2024)Bogyrbayeva, Meraliyev, Mustakhov, and Dauletbayev}]{bogyrbayeva2024machine}
Bogyrbayeva, Aigerim, Meraryslan Meraliyev, Taukekhan Mustakhov, Bissenbay Dauletbayev. 2024.
\newblock Machine learning to solve vehicle routing problems: A survey.
\newblock {\it IEEE Transactions on Intelligent Transportation Systems\/}, { 25} (6), 4754-4772.

\bibitem[{Brunetta et~al.(1997)Brunetta, Conforti, and Rinaldi}]{brunetta1997branch}
Brunetta, Lorenzo, Michele Conforti, Giovanni Rinaldi. 1997.
\newblock A branch-and-cut algorithm for the equicut problem.
\newblock {\it Mathematical Programming\/}, { 78} (2), 243-263.

\bibitem[{Cao et~al.(2025)Cao, Wang, and Xiong}]{Cao2025}
Cao, Linjiang, Maonan Wang, Xi~Xiong. 2025.
\newblock A large language model-enhanced q-learning for capacitated vehicle routing problem with time windows.
\newblock {\it arXiv preprint arXiv:2505.06178\/}, .

\bibitem[{Cleophas and Ehmke(2014)}]{cleophas2014deliveries}
Cleophas, Catherine, Jan~Fabian Ehmke. 2014.
\newblock When are deliveries profitable? considering order value and transport capacity in demand fulfillment for last-mile deliveries in metropolitan areas.
\newblock {\it Business \& Information Systems Engineering\/}, { 6} (3), 153-163.

\bibitem[{Dantzig and Ramser(1959)}]{dantzig1959vrp}
Dantzig, George~B., John~H. Ramser. 1959.
\newblock The truck dispatching problem.
\newblock {\it Management Science\/}, { 6} (1), 80-91.

\bibitem[{Dhara and Barba(2023)}]{Dhara2023}
Dhara, Sasank, Sebastian~Delgado Barba. 2023.
\newblock Large language models in supply chain management, .

\bibitem[{Ehrler et~al.(2021)Ehrler, Sch{\"o}der, and Seidel}]{ehrler2021challenges}
Ehrler, Verena~Ch, Dustin Sch{\"o}der, Saskia Seidel. 2021.
\newblock Challenges and perspectives for the use of electric vehicles for last mile logistics of grocery e-commerce--findings from case studies in germany.
\newblock {\it Research in Transportation Economics\/}, { 87} 100757.

\bibitem[{Furnon and Perron(2010)}]{ortools_routing}
Furnon, Vincent, Laurent Perron. 2010.
\newblock Or-tools routing library.
\newblock \urlprefix\url{https://developers.google.com/optimization/routing/}.

\bibitem[{Gou et~al.(2021)Gou, Yu, Maybank, and Tao}]{gou2021knowledge}
Gou, Jianping, Baosheng Yu, Stephen~J Maybank, Dacheng Tao. 2021.
\newblock Knowledge distillation: A survey.
\newblock {\it International journal of computer vision\/}, { 129} (6), 1789-1819.

\bibitem[{Huang et~al.(2024{\natexlab{a}})Huang, Liu, Fu, Wu, Mukadam, Malik, Goldberg, and Abbeel}]{huang2024early}
Huang, Huang, Fangchen Liu, Letian Fu, Tingfan Wu, Mustafa Mukadam, Jitendra Malik, Ken Goldberg, Pieter Abbeel. 2024{\natexlab{a}}.
\newblock Early fusion helps vision language action models generalize better, .

\bibitem[{Huang et~al.(2025)Huang, Zhang, Feng, Wu, and Tan}]{huang2025multimodal}
Huang, Yuxiao, Wenjie Zhang, Liang Feng, Xingyu Wu, Kay~Chen Tan. 2025.
\newblock How multimodal integration boost the performance of llm for optimization: Case study on capacitated vehicle routing problems.
\newblock {\it 2025 IEEE Symposium for Multidisciplinary Computational Intelligence Incubators (MCII)\/}. IEEE, 1-7.

\bibitem[{Huang et~al.(2024{\natexlab{b}})Huang, Shi, and Sukhatme}]{Huang2024}
Huang, Zhehui, Guangyao Shi, Gaurav~S Sukhatme. 2024{\natexlab{b}}.
\newblock From words to routes: Applying large language models to vehicle routing.
\newblock {\it CoRR\/}, .

\bibitem[{IBEF(2025)}]{IBEF2025}
IBEF. 2025.
\newblock E-commerce industry in india.
\newblock Tech. rep., India Brand Equity Foundation.

\bibitem[{Jannelli et~al.(2024)Jannelli, Schoepf, Bickel, Netland, and Brintrup}]{Jannelli2024}
Jannelli, Valeria, Stefan Schoepf, Matthias Bickel, Torbjørn Netland, Alexandra Brintrup. 2024.
\newblock Agentic llms in the supply chain: Towards autonomous multi-agent consensus-seeking.
\newblock {\it arXiv preprint arXiv:2411.10184\/}, .

\bibitem[{Jiang et~al.(2024)Jiang, Sablayrolles, Roux, Mensch, Savary, Bamford, Chaplot, Casas, Hanna, Bressand et~al.}]{jiang2024mixtral}
Jiang, Albert~Q, Alexandre Sablayrolles, Antoine Roux, Arthur Mensch, Blanche Savary, Chris Bamford, Devendra~Singh Chaplot, Diego de~las Casas, Emma~Bou Hanna, Florian Bressand, et~al. 2024.
\newblock Mixtral of experts.
\newblock {\it arXiv preprint arXiv:2401.04088\/}, .

\bibitem[{Jiang et~al.(2025)Jiang, Wu, Zhang, and Zhang}]{Jiang2025}
Jiang, Xia, Yaoxin Wu, Chenhao Zhang, Yingqian Zhang. 2025.
\newblock Droc: Elevating large language models for complex vehicle routing via decomposed retrieval of constraints.
\newblock {\it 13th international Conference on Learning Representations, ICLR 2025\/}.

\bibitem[{Kikuta et~al.(2024)Kikuta, Ikeuchi, Tajiri, and Nakano}]{Kikuta2024}
Kikuta, Daisuke, Hiroki Ikeuchi, Kengo Tajiri, Yuusuke Nakano. 2024.
\newblock Routeexplainer: an explanation framework for vehicle routing problem.
\newblock {\it Pacific-asia conference on knowledge discovery and data mining\/}. 30-42.

\bibitem[{Labadie et~al.(2016)Labadie, Prins, and Prodhon}]{Labadie2016}
Labadie, Nacima, Christian Prins, Caroline Prodhon. 2016.
\newblock {\it Metaheuristics for Vehicle Routing Problems\/}, vol.~3.
\newblock \doi{10.1002/9781119136767}.

\bibitem[{Li et~al.(2024)Li, Jiang, Gadepally, and Tiwari}]{li2024llm}
Li, Baolin, Yankai Jiang, Vijay Gadepally, Devesh Tiwari. 2024.
\newblock Llm inference serving: Survey of recent advances and opportunities.
\newblock {\it 2024 IEEE High Performance Extreme Computing Conference (HPEC)\/}. IEEE, 1-8.

\bibitem[{Li et~al.(2023)Li, Mellou, Zhang, Pathuri, and Menache}]{Li2023}
Li, Beibin, Konstantina Mellou, Bo~Zhang, Jeevan Pathuri, Ishai Menache. 2023.
\newblock Large language models for supply chain optimization.
\newblock {\it arXiv preprint arXiv:2307.03875\/}, .

\bibitem[{Li et~al.(2025)Li, Zheng, Hao, and Wang}]{Li2025b}
Li, Kai, Ruihao Zheng, Xinye Hao, Zhenkun Wang. 2025.
\newblock Multi-objective infeasibility diagnosis for routing problems using large language models.
\newblock {\it arXiv preprint arXiv:2508.03406\/}, .

\bibitem[{Liao et~al.(2021)Liao, Kar, and Fidler}]{liao2021towards}
Liao, Yuan-Hong, Amlan Kar, Sanja Fidler. 2021.
\newblock Towards good practices for efficiently annotating large-scale image classification datasets.
\newblock {\it Proceedings of the IEEE/CVF Conference on Computer Vision and Pattern Recognition\/}. 4350-4359.

\bibitem[{Liu et~al.(2023{\natexlab{a}})Liu, Lu, Gui, Zhang, Tong, and Yuan}]{Liu2023a}
Liu, Fei, Chengyu Lu, Lin Gui, Qingfu Zhang, Xialiang Tong, Mingxuan Yuan. 2023{\natexlab{a}}.
\newblock Heuristics for vehicle routing problem: A survey and recent advances.
\newblock {\it arXiv preprint arXiv:2303.04147\/}, .

\bibitem[{Liu et~al.(2023{\natexlab{b}})Liu, Wu, Liu, Wang, Wang, and Qu}]{Liu2023b}
Liu, Yang, Fanyou Wu, Zhiyuan Liu, Kai Wang, Feiyue Wang, Xiaobo Qu. 2023{\natexlab{b}}.
\newblock Can language models be used for real-world urban-delivery route optimization?
\newblock {\it The Innovation\/}, { 4}.

\bibitem[{Ma et~al.(2025)Ma, Peng, Yang, Shen, Koedinger, and Wu}]{ma2025should}
Ma, Qianou, Weirui Peng, Chenyang Yang, Hua Shen, Ken Koedinger, Tongshuang Wu. 2025.
\newblock What should we engineer in prompts? training humans in requirement-driven llm use.
\newblock {\it ACM Transactions on Computer-Human Interaction\/}, { 32} (4), 1-27.

\bibitem[{Macioszek(2017)}]{macioszek2017first}
Macioszek, El{\.z}bieta. 2017.
\newblock First and last mile delivery--problems and issues.
\newblock {\it Scientific and technical conference transport systems theory and practice\/}. Springer, 147-154.

\bibitem[{Maersk(2025)}]{Maersk2025}
Maersk. 2025.
\newblock Closing the logistics loop with last-mile delivery.

\bibitem[{Marcondes et~al.(2025)Marcondes, Gala, Magalh{\~a}es, Perez~de Britto, Dur{\~a}es, and Novais}]{marcondes2025using}
Marcondes, Francisco~S, Adelino Gala, Renata Magalh{\~a}es, Fernando Perez~de Britto, Dalila Dur{\~a}es, Paulo Novais. 2025.
\newblock Using ollama.
\newblock {\it Natural Language Analytics with Generative Large-Language Models: A Practical Approach with Ollama and Open-Source LLMs\/}. Springer, 23-35.

\bibitem[{Masry et~al.(2022)Masry, Long, Tan, Joty, and Hoque}]{masry2022chartqabenchmarkquestionanswering}
Masry, Ahmed, Do~Xuan Long, Jia~Qing Tan, Shafiq Joty, Enamul Hoque. 2022.
\newblock Chartqa: A benchmark for question answering about charts with visual and logical reasoning.
\newblock \urlprefix\url{https://arxiv.org/abs/2203.10244}.

\bibitem[{Panigrahi et~al.(2016)Panigrahi, Upadhyaya, Raichurkar et~al.}]{panigrahi2016commerce}
Panigrahi, CMA, Ranjan Upadhyaya, PP~Raichurkar, et~al. 2016.
\newblock E-commerce services in india: prospects and problems.
\newblock {\it International Journal on Textile Engineering and Processes\/}, { 2} (1),.

\bibitem[{Puram et~al.(2022)Puram, Gurumurthy, Narmetta, and Mor}]{puram2022last}
Puram, Praveen, Anand Gurumurthy, Mukesh Narmetta, Rahul~S Mor. 2022.
\newblock Last-mile challenges in on-demand food delivery during covid-19: understanding the riders' perspective using a grounded theory approach.
\newblock {\it The International Journal of Logistics Management\/}, { 33} (3), 901-925.

\bibitem[{Ramachandran et~al.(2019)Ramachandran, Parmar, Vaswani, Bello, Levskaya, and Shlens}]{ramachandran2019stand}
Ramachandran, Prajit, Niki Parmar, Ashish Vaswani, Irwan Bello, Anselm Levskaya, Jon Shlens. 2019.
\newblock Stand-alone self-attention in vision models.
\newblock {\it Advances in neural information processing systems\/}, { 32}.

\bibitem[{Raman(2021)}]{raman2021comprehensive}
Raman, Pradeep~Kannan. 2021.
\newblock Comprehensive analysis of ecommerce and marketplaces: Global perspectives with emphasis on the indian context.
\newblock {\it Asian Journal of Multidisciplinary Research \& Review\/}, { 2} (1), 1-52.

\bibitem[{Shi et~al.(2024)Shi, Wu, Mao, Wang, and Darrell}]{shi2024we}
Shi, Baifeng, Ziyang Wu, Maolin Mao, Xin Wang, Trevor Darrell. 2024.
\newblock When do we not need larger vision models?
\newblock {\it European Conference on Computer Vision\/}. Springer, 444-462.

\bibitem[{Shi et~al.(2019)Shi, Sun, Teng, and Hu}]{Shi2019}
Shi, Haiyang, Lijun Sun, Yue Teng, Xiangpei Hu. 2019.
\newblock An online intelligent vehicle routing and scheduling approach for b2c e-commerce urban logistics distribution.
\newblock {\it Procedia computer science\/}, { 159} 2533-2542.

\bibitem[{Sidorov and Morozov(2021)}]{Sidorov2021}
Sidorov, Konstantin, Alexander Morozov. 2021.
\newblock A review of approaches to modeling applied vehicle routing problems.
\newblock {\it arXiv preprint arXiv:2105.10950\/}, .

\bibitem[{Toth and Vigo(2002)}]{Toth2002}
Toth, Paolo, Daniele Vigo. 2002.
\newblock {\it The vehicle routing problem\/}.
\newblock SIAM.

\bibitem[{Tran et~al.(2025)Tran, Nguyen-Tri, Binh, and Thanh-Tung}]{Thanh2025}
Tran, Cong~Dao, Quan Nguyen-Tri, Huynh Thi~Thanh Binh, Hoang Thanh-Tung. 2025.
\newblock Large language models powered neural solvers for generalized vehicle routing problems.
\newblock {\it Towards Agentic AI for Science: Hypothesis Generation, Comprehension, Quantification, and Validation\/}.

\bibitem[{Ulmer et~al.(2017)Ulmer, Goodson, Mattfeld, and Thomas}]{Ulmer2017}
Ulmer, Marlin~W, Justin~C Goodson, Dirk~C Mattfeld, Barrett~W Thomas. 2017.
\newblock Dynamic vehicle routing: Literature review and modeling framework, .

\bibitem[{Wu et~al.(2024)Wu, Wang, Wen, Xiao, Wu, Wu, Yu, Maskell, and Zhou}]{Wu2024}
Wu, Xuan, Di~Wang, Lijie Wen, Yubin Xiao, Chunguo Wu, Yuesong Wu, Chaoyu Yu, Douglas~L Maskell, You Zhou. 2024.
\newblock Neural combinatorial optimization algorithms for solving vehicle routing problems: A comprehensive survey with perspectives.
\newblock {\it arXiv preprint arXiv:2406.00415\/}, .

\bibitem[{Xiao and Zhu(2025)}]{xiao2025foundations}
Xiao, Tong, Jingbo Zhu. 2025.
\newblock Foundations of large language models.
\newblock {\it arXiv preprint arXiv:2501.09223\/}, .

\bibitem[{Yang et~al.(2023)Yang, Wang, Lu, Liu, Le, Zhou, and Chen}]{Yang2023}
Yang, Chengrun, Xuezhi Wang, Yifeng Lu, Hanxiao Liu, Quoc~V Le, Denny Zhou, Xinyun Chen. 2023.
\newblock Large language models as optimizers.
\newblock {\it The Twelfth International Conference on Learning Representations\/}.

\bibitem[{Ye et~al.(2024)Ye, Wang, Cao, Berto, Hua, Kim, Park, and Song}]{Ye2024}
Ye, Haoran, Jiarui Wang, Zhiguang Cao, Federico Berto, Chuanbo Hua, Haeyeon Kim, Jinkyoo Park, Guojie Song. 2024.
\newblock Reevo: Large language models as hyper-heuristics with reflective evolution.
\newblock {\it Advances in neural information processing systems\/}, { 37} 43571-43608.

\bibitem[{Zhai et~al.(2022)Zhai, Kolesnikov, Houlsby, and Beyer}]{zhai2022scaling}
Zhai, Xiaohua, Alexander Kolesnikov, Neil Houlsby, Lucas Beyer. 2022.
\newblock Scaling vision transformers.
\newblock {\it Proceedings of the IEEE/CVF conference on computer vision and pattern recognition\/}. 12104-12113.

\bibitem[{Zhou et~al.(2024{\natexlab{a}})Zhou, Cao, Wu, Song, Ma, Zhang, and Xu}]{Zhou2024b}
Zhou, Jianan, Zhiguang Cao, Yaoxin Wu, Wen Song, Yining Ma, Jie Zhang, Chi Xu. 2024{\natexlab{a}}.
\newblock Mvmoe: Multi-task vehicle routing solver with mixture-of-experts.
\newblock {\it arXiv preprint arXiv:2405.01029\/}, .

\bibitem[{Zhou et~al.(2024{\natexlab{b}})Zhou, Zhou, and Liu}]{Zhou2024}
Zhou, Ziai, Bin Zhou, Hao Liu. 2024{\natexlab{b}}.
\newblock Dynamicroutegpt: A real-time multi-vehicle dynamic navigation framework based on large language models.
\newblock {\it arXiv preprint arXiv:2408.14185\/}, .

\end{thebibliography}


\begin{thebibliography}{}


    \bibitem[{Doe et~al.(2006)}]{Doe2014}
    Doe, F., J. Smith, G. Toledano, R. Richard. 2006.
    Title of the article here. {\it Journal of Science,} 89 (5), 882--887.

        \bibitem[{Smith et~al.(2016)}]{Doe2009}
Smith, A., J. Roe, R. Doe.  2016.
    Another article title. {\it Another Journal Name,} 9 (1), 85--87.

\end{thebibliography}

\end{document}